\documentclass{article}
\usepackage[utf8]{inputenc}

\title{Variational Inference for Policy Gradient}
\author{Tianbing Xu}
\date{February 2018}

\usepackage{cite}
\usepackage{graphicx}
\usepackage{qiangstyle}
\usepackage[toc,page]{appendix}

\newtheorem{lemma}{Lemma}

\newcommand{\pian}[2]{\frac{\partial #1}{\partial #2}}

\begin{document}

\maketitle

\begin{abstract}
Inspired by the seminal work on Stein Variational Inference \cite{SVGD} and Stein Variational Policy Gradient \cite{SVPG}, we derived a method to generate samples from the posterior variational parameter distribution by \textit{explicitly} minimizing the KL divergence to match the target distribution in an amortize fashion. Consequently, we applied this varational inference technique into vanilla policy gradient, TRPO and PPO with Bayesian Neural Network  parameterizations for reinforcement learning problems.
\end{abstract}

\section{Parametric Minimization of KL Divergence}
\label{SecMinKL}
Suppose we have a random sample from a base distribution $\xi\sim q_0(\xi)$, 
e.g. $q_0 = \mathcal{N}(0, \mathcal{I})$, we are able to generate an induced distribution $q_\phi(\theta)$ 
by the general \textbf{invertible} and differentiable transformation $\theta=h_\phi(\xi)$ (see Appendix A).  Our goal is to 
regard $q_\phi(\theta)$ as a variational distribution to match the true distribution $p(\theta)$ such that $J=KL(q_\phi(\theta)||p(\theta))$ is minimized.

\begin{lemma}
\begin{equation}
H(q) = H(q_0) + E_{\xi \sim q_0}\left(\log\det\left(\pian{h_\phi(\xi)}{\xi}\right)\right) \label{EqDiffEntropy}
\end{equation} 
\end{lemma}
    
with (\ref{EqDiffEntropy}), we can have the following identity for $KL(q_\phi(\theta)||p(\theta))$ : 
\begin{eqnarray*}
KL(q||p) &=& -H(q) - E_{q(\theta)}(\log p(\theta)) \\
        &=& -H(q_0)-E_{\xi\sim q_0}\left(\log\det\left(\pian{h_\phi(\xi)}{\xi}\right)\right) - 
        E_{\xi \sim q_0}(\log p(h_\phi(\xi))) \\
        &=& -H(q_0) - E_{\xi \sim q_0}\left(\log\det\left(\pian{h_\phi(\xi)}{\xi}\right) + \log p(h_\phi(\xi))\right)
\end{eqnarray*}
    
Hence the gradient of $KL(q||p)$ with respect to the parameters of transformation mapping $\phi$ is:
\begin{equation}
\pian{KL}{\phi} = -E_{\xi \sim q_0}\left[\pian{\log p(h_\phi(\xi))}{\phi} + \pian{}{\phi}\log\det\left(\pian{h_\phi(\xi)}{\xi}\right)\right] \label{EqGrad}
\end{equation}
    
Note that the first term is the usual log-likelihood term, and the second term serves as a repulsive force preventing all $\xi$'s from collapsing towards the maximum likelihood estimation. We can perform stochastic gradient descent using (\ref{EqGrad}) to find the optimal $\phi$. This method is related to the interesting and seminal Stein Variational Inference \cite{AmortizedSVGD}, the major difference is that the later one uses kernelized Stein variational gradient, while we use log determinate as the repulsive force.

\section{Bayesian Formulation of Variational RL}
\label{SecBayesianRL}
We generate the policy distribution from Bayesian Neural Network. Suppose $\theta$ is the parameter of 
the policy network, the parameter is able to generated from a base distribution 
$\xi \sim \mathcal{N}(0, \mathcal{I})$ with an \textbf{invertible} and differentiable transformation function $h_\phi(\xi)$.
We may adopt complicated differentiable transformation functions $h_\phi(\xi)$. For a simple example, the 
weight for each connection of neuron could be $\theta_i = \mu_i + \sigma_i$. 
For each realized weight parameter $\theta$, we are able to generate a stochastic multi-modal policy distribution $\pi_\theta(a | s)$ 
represented as a Neural Network with several hidden layers.

$R(\theta)$ is the expected cumulative reward under policy $\pi_\theta$,
\begin{eqnarray*}
R(\theta) = E_{\pi_{\theta}} \left[ \sum_t {\gamma^{t} r(s_t, a_t)} \right]
\end{eqnarray*}

$P$ is a distribution over $\theta$ and $H(P)$ is the Shannon entropy of $P$. We want to find $P$ to maximize the following objective:
\begin{eqnarray*}
    \tilde{R} = \int R(\theta) dP(\theta) + \alpha H(P)
\end{eqnarray*}
    
\begin{eqnarray*}
\tilde{R} &=& \alpha \int \left( \log\left(\frac{1}{p(\theta)}\right)+\frac{1}{\alpha}R(\theta)\right) dP(\theta) \\
        &=& \int \log\left(\frac{\exp(\frac{1}{\alpha}R(\theta))}{p(\theta)}\right) dP(\theta)
\end{eqnarray*}
The optimal $P$ is:
\begin{align}
\label{eq:posterior}
p(\theta) \propto \exp\left(\frac{1}{\alpha}R(\theta)\right)
\end{align}
    
This formulation is originally proposed in \cite{SVPG}. The difficulty of this formulation is calculating the 
normalization factor $\int \exp(\frac{1}{\alpha} R(\theta)) d\theta$. We are able to bypass it by calculating the gradient of the its log-probability which was a exciting idea from Stein variational inference\cite{SVGD} and similarly, here we can use Eq.~\ref{EqGrad}. 
Suppose we generate sample of $\theta$ by transforming random noise $\xi$ using 
$h_\phi(\xi)$. Let $q_\phi(\theta)$ be the induced variational distribution from the transformation. 
Our optimization objective is to match the induced variational distribution and the 'true' policy parameter distribution by minimizing $KL(q||p)$. The gradients for the parameters ($\phi$) of policy distribution are,
\begin{equation}
\label{eq:grad}
-\pian{KL}{\phi} = E_{\xi \sim q_0}\left[\frac{1}{\alpha}\pian{R(\theta)}{\theta} \pian{\theta}{\phi} 
        + \pian{}{\phi}\log\det\left(\pian{h_\phi(\xi)}{\xi}\right)\right]
\end{equation}
where $\pian{R}{\theta}$ can be calculated using the standard policy gradient formula,
\begin{equation}
\label{eq:pg}
\pian{R(\theta)}{\theta} = E_{\pi_{\theta}}\left[\sum_{t} \pian{\log\pi_{\theta}(a_t|s_t)}{\theta}A(s_t, a_t)\right]
\end{equation}
We can sample different $\theta$ for exploration in different sessions. We may want to decrease $\alpha$ during the training to anneal the temperature to stationary parameter distributions.

$A(s_t, a_t)$ is the advantage function, it can be estimated as $A(s_t, a_t) = Q(s_t, a_t)- b(s_t)$ 
or let baseline $b(s_t) = V(s_t)$, or $A(s_t, a_t) = r(s_t, a_t) + V(s_t) - V(s_{t+1})$, 
and $Q(s_t, a_t)$ is the state-action $Q$ value and $V(s_t)$ is the value function. For more sophisticated estimation, we can use GAE (Generalized Advantage Estimation, \cite{GAE}).

\textbf{Example 1:}
For a simple transformation of the following form:
\begin{equation*}
    \theta_i = \mu_i + \sigma_i \xi_i
\end{equation*}
the sample gradient estimation w.r.t $\mu_i$ and $\sigma_i$ is:
\begin{eqnarray*}
    -\pian{KL}{\mu_i} &=& \frac{1}{\alpha}\pian{R(\theta)}{\theta_i} \\
    -\pian{KL}{\sigma_i} &=& \frac{1}{\alpha}\xi_i\pian{R(\theta)}{\theta_i} + \frac{1}{\sigma_i} 
\end{eqnarray*}
    
\section{Variational Policy Gradient with Transformation}
We introduce our varational inference into vanilla policy gradient REINFORCE \cite{VPG}.
Given a realization of network parameter $\theta$, in order to generate a stochastic policy distribution, we introduce another random noise $\zeta \sim \pi_0(\cdot)$. With the second \textbf{invertible} and differentiable transformation $a = g_\theta(s, \zeta)$, it induces a stochastic policy distribution $a \sim \pi_\theta(a | s)$ in the closed form, 
\begin{eqnarray}
\pi_\theta(a_t |s_t) = \frac{\pi_0(g^{-1}_\theta(a,s))}{\det\left(\pian{g_\theta(s,\zeta)}{\zeta}\right)}
\end{eqnarray}
Hence, the policy gradient is,
\begin{align}
\pian{R(\theta)}{\theta} = \nonumber \\
E_{\pi_{\theta}} \left\{
\sum_{t} \left[\pian{}{\theta} \log \pi_0(g^{-1}_\theta(a_t,s_t)) - 
\pian{}{\theta}\log \det\left(\pian{g_\theta(s_t,\zeta)}
{\zeta}\right) \right] A(s_t,a_t) 
\right\}
\end{align}

When the inverse of transformation $\zeta = g^{-1}_\theta(a, s)$ is 
difficult to calculate, we could use $g_\theta(s, \zeta)$ directly,
\begin{align*}
\pian{R(\theta)}{\theta} = \nonumber \\
E_{\pi_{\theta}} \left\{
\sum_t \left[ \pian{}{\zeta} \log \pi_0(\zeta) \pian {\zeta}{g_\theta(s_t, \zeta)} 
\pian{g_\theta(s_t, \zeta)} {\theta}
- \pian{}{\theta} 
\log\det\left(\pian{g_\theta(s_t,\zeta)}{\zeta}\right) \right] 
A(s_t, a_t) \right\}
\end{align*}

\subsection{Simple Policy Network Parameterization}
\label{sec:gen}
We adopt a very simple yet general representative generative model. The policy parameter is
generated from noise $\xi$ with transformation $h_\phi(s, \xi)$, which is a neural network parameterized with $\phi$. With another noise $\zeta$, we generate the action $a$ by another transformation $g_\theta(s, \zeta)$, parameterized with $\theta$, from policy network distribution.
\begin{eqnarray*}
 \xi \sim N(0, \mathcal{I}), \zeta \sim N(0, \mathcal{I})\\
 \theta = h_{\phi}(s, \xi) = \mu_\phi(s) + \xi \cdot \sigma_\phi(s)\\
 a = g_\theta(s, \zeta) = \theta(s, \xi) + \zeta  \\
\end{eqnarray*}
This induces a simple policy distribution,
\begin{align*}
    \pi_\theta(a | s, \theta) \propto \exp \left( -0.5 * (a - \theta)^T (a - \theta) \right)
\end{align*}

$\mu_\phi(s)$ and $\sigma_\phi(s)$ are mean and variance networks, with weight parameter $\phi$, they are used to generate the posterior distribution of action mean $\theta$. For continuous control problems, we use MLP (multilayer perceptron) to represents the mean and variance networks. Then we can find the variational policy parameter distribution by minimizing the KL divergence between the variation distribution $q_\phi(\theta)$ generated based on the transformation $h_\phi(s, \xi)$ and the optimal posterior parameter distribution $p(\theta)$ (energy-based model, Eq.~\ref{eq:posterior}) as $KL(q_\phi(\theta) || \exp \{ R(\theta)\})$.

From the full complete gradient of Eq.(\ref{eq:grad}), we have,
\begin{align}
    -\pian{KL}{\phi} = E_{\xi \sim q_0}\left[
     \frac{1}{\alpha}\pian{R(\theta)}{\theta} \pian{h_{\phi}(s, \xi)}{\phi} +  {\sum_{i = 0}^d} \pian{\log \sigma_{\phi}^{i}(s)} {\phi}
    \right]
\end{align}
It is straightforward to auto diff $\pian{R(\phi)}{\phi}$ and $\pian{\log \sigma_{\phi}(s)} {\phi}$.
$\pian{\sigma_{\phi}(s)} {\phi}$ is the backprop of variance network $\sigma_{\phi}$, for the simplest example, let $\sigma_\phi(s) = \sigma(w^T s)$ a sigmoid function, we have,
\begin{align*}
    \sum_{i = 0}^d \frac{1}{\sigma_{\phi}^{i}(s)} \pian{\sigma_{\phi}^{i}(s)} {\phi} =
    \sum_{i = 0}^d  (1 - \sigma_\phi^i (w^T s)) s
\end{align*}    

\subsection{Auxiliary Policy Network Parameterization}
For a more general parameterization of the policy, instead of regarding $\theta$ as a parameter of policy, we can take $\theta$ as a random variable, and introduce the  auxilliary network parameter $\Psi$, then the action $a$ is generated from noise $\zeta$ by transformation $g_\Psi(s, \theta, \zeta)$, it induces the corresponding policy distribution $\pi(a | s, \theta, \Psi)$. An example of $g_\Psi(s, \theta, \zeta)$ could be a MLP as,
\begin{align*}
    g_\Psi(s, \theta, \zeta) = MLP_\Psi(\theta(s,\xi), s) + \zeta
\end{align*}
    
Similarly, the posterior of $\theta$ is,
\begin{align*}
    p(\theta) \propto p_0(\theta) \exp \{\frac{1}{\alpha} R_\Psi(\theta)\}
\end{align*}
here the cumulative rewards,
\begin{align*}
    R_\Psi(\theta) = E_{\pi(\theta, \Psi)} \left[ \sum_{t} \gamma^{t} r(s_t, a_t) \right]
\end{align*}
This gives us a more general representation of the policy, compared to the previous formulation in Section~\ref{sec:gen}. Furthermore, it is easy to introduce multi-modal distribution for the stochastic actions.
    
The gradient of KL divergence between the variational distribution and posterior $p(\theta)$ is Eq.~\ref{eq:grad}.
In addition, we need to learn network parameter $\Psi$,
\begin{align*}
    \pian{R_\Psi(\theta)} {\Psi} = E_{\pi(\theta, \Psi)} 
    \left[ \sum_{t} 
    \pian{\log \pi_{(\theta, \Psi)}(a_t |s_t)} {\Psi} A(s_t, a_t)
    \right]
\end{align*}    
    
\section{Connection to TRPO}
The motivation is to combine fast convergence with sample efficiency of TRPO \cite{TRPO} and the exploration introduced by variational inference of posterior policy parameter distribution.

The TRPO objective,
\begin{align*}
L(\theta) = E_{\theta_{old}} \left[ \frac{\pi_\theta(a|s)}{\pi_{old}(a|s)} 
    A^{\pi_{old}}(a|s)\right]  \\
s.t. \quad D_{KL}(\pi_\theta, \pi_{old}) \leq \delta
\end{align*}

The TRPO Variational Policy Gradient,
\begin{align}
-\pian{KL}{\phi} 
= E_{\xi \sim q_0}\left[\frac{1}{\alpha}\pian{L(\theta)}{\theta} \pian{\theta}{\phi} 
        + \pian{}{\phi}\log\det(\pian{h_\phi(\xi)}{\xi})\right]
\end{align}

TRPO Variational Policy Update:
\begin{align*}
\phi \gets \phi - \eta H^{-1}(\theta) \nabla_\phi KL 
\end{align*}

Another important point is how to calculate the $KL$ divergence between the current and previous policies efficiently?
\begin{align}
   D_{KL}(\pi_\theta(\cdot|s, \pi_{\theta_{old}}(\cdot | s))) = KL(\mu_\theta(s), \mu_{\theta_{old}}(s)) \\ \nonumber
   \approx \frac{1}{2} (\theta - \theta_{old})^T H(\theta_{old}) (\theta - \theta_{old}) 
\end{align}   
   
To get the Fisher information matrix, first method is to compute the Hessian of averaged $KL$ divergence,
\begin{align}
   H(\theta_{old})_{i,j} = E_s \left[ \frac{\partial^2 } {\partial \theta_i \partial \theta_j} 
   D_{KL}(\pi_\theta(\cdot|s, \pi_{\theta_{old}}(\cdot | s))) \right]
\end{align}
This is equivalent to calculate the second derivative of $KL(\mu_\theta(s), \mu_{\theta_{old}})$ w.r.t. $\theta$.
   
The other method is using Covariance matrix,
\begin{align}
    H(\theta_{old}) = E_s \left[ \left(\pian{} {\theta} {\log \pi_\theta(\cdot | s)} \right)
    \left(\pian{} {\theta} {\log \pi_\theta(\cdot | s)} \right)^T \right]   
\end{align}

\section{Connection to PPO}
It is natural to adopt our variational inference method to PPO (Proximal Policy Optimization) \cite{PPO}.
The objective function of PPO to be maximized is,
\begin{align*}
J_{ppo} (\theta) = E_{old} [\frac{\pi_\theta(a|s)}{\pi_{old}(a|s)} Q^{\pi}(a|s) - 
    \lambda KL(\pi_{old}, \pi_\theta)]
\end{align*}
where $KL(\pi_{old}, \pi_{\theta})= E_{\pi_{old}}\left[ KL(\pi_{old}(\cdot | s), \pi_\theta(\cdot | s)) \right]$
    
PPO Variational Policy Gradient,
\begin{align*}
-\pian{KL}{\phi} 
= E_{\xi \sim q_0}\left[\frac{1}{\alpha}\pian{J_{ppo}(\theta)}{\theta} \pian{\theta}{\phi} 
        + \pian{}{\phi}\log\det(\pian{h_\phi(\xi)}{\xi})\right]
\end{align*}

The PPO Variational Policy Update is, 
\begin{align*}
\phi \gets \phi -\eta \nabla_\phi KL 
\end{align*}

\appendix
%\begin{appendices}
\section{random variable transformation}
Given random variable $X$, we introduce the transformation function $Y = f(X)$ to generate random variable $Y$. The transformation function, 
\begin{align*}
    f : \mathcal{R} \rightarrow \mathcal{R}
\end{align*}
$f$ needs to be \textbf{invertible}, the inverse image of $f$ of set $A$,
\begin{align*}
    f^{-1}(A) = \{ x \in \mathcal{R}, f(x) \in A \}
\end{align*}
The inverse image also needs to satisfied the following requirements.
\begin{enumerate}
    \item {$f^{-1} (\mathcal{R}) = \mathcal{R}$ }
    \item {$f^{-1} (A^c) = {f^{-1}(A)}^c$}
    \item {$f^{-1}(\bigcup_\lambda A_\lambda) = \bigcup_\lambda f^{-1}(A_\lambda)$, for any sets \{$A_\lambda$, $\lambda \in \Omega$ \}
    }  
\end{enumerate}
    
Assume the distribution of r.v. $X$ and $Y$ are $p_X(x)$ and $p_Y(y)$, we have, 
\begin{align}
    p_Y(y) = p_X(f^{-1}(y)) \det \left(\pian{f^{-1}(y)} {y} \right)
\end{align}
    
%\end{appendices}

\bibliographystyle{plain}
\bibliography{vpg}
\end{document}